\documentclass[10pt,twocolumn,letterpaper]{article}
\pdfoutput=1

\usepackage{cvpr}
\usepackage{times}
\usepackage{epsfig}
\usepackage{graphicx}
\usepackage{amsmath}
\usepackage{amssymb}
\usepackage{color}
\usepackage{array}
\usepackage{multirow}
\newcolumntype{M}[1]{>{\centering\arraybackslash}m{#1}}
\newcolumntype{P}[1]{>{\centering\arraybackslash}p{#1}}

\definecolor{mygray}{rgb}{0.5, 0.5, 0.5}
\usepackage[pagebackref=true,breaklinks=true,letterpaper=true,colorlinks,bookmarks=false]{hyperref}

\cvprfinalcopy % *** Uncomment this line for the final submission

\def\cvprPaperID{1112} % *** Enter the CVPR Paper ID here

\ifcvprfinal\fi
\begin{document}

%%%%%%%%% TITLE
\title{Object Detection Free Instance Segmentation With Labeling Transformations}

\author{Long Jin$^1$, Zeyu Chen$^1$,  Zhuowen Tu$^{2,1}$\\
$^1$Dept. of CSE and $^2$Dept. of CogSci, University of California, San Diego\\
9500 Gilman Drive, La Jolla, CA 92093 \\
{\tt\small \{longjin, zec003, ztu\}@ucsd.edu}
%\and
}

\maketitle
%\thispagestyle{empty}

%%%%%%%%% ABSTRACT
\begin{abstract}
%Existing methods for instance image segmentation mostly follow a standard pipeline including (1) proposal generation, (2) object detection, and (3) instance masking, which is a detour from the basic image segmentation task in which pixel-wise labeling is directly performed.
Instance segmentation has attracted recent attention in computer vision and existing methods in this domain mostly have an object detection stage. In this paper, we study the intrinsic challenge of the instance segmentation problem, the presence of a quotient space (swapping the labels of different instances leads to the same result), and propose new methods that are object proposal- and object detection- free. We propose three alternative methods, namely pixel-based affinity mapping, superpixel-based affinity learning, and boundary-based component segmentation, all focusing on performing labeling transformations to cope with the quotient space problem. By adopting fully convolutional neural networks (FCN) like models, our framework attains competitive results on both the PASCAL dataset (object-centric) and the Gland dataset (texture-centric), which the existing methods are not able to do. Our work also has the advantages in its transparency, simplicity, and being all segmentation based.

\end{abstract}

%----------------------------------------------------------------------------
%%%%%%%%% BODY TEXT
\section{Introduction}

Object detection and semantic segmentation are both important tasks in computer vision. The goal of object detection is to predict the bounding box as well as the semantic class of each object, whereas semantic segmentation focuses on predicting the semantic class of the individual pixels in an image. In general, object detection does not provide accurate pixel-level object segmentation and semantic segmentation ignores to distinguish different objects in the same class. 

\begin{figure}[!htp]
\begin{center}
\begin{tabular} {c}
\includegraphics[width=0.9\linewidth]{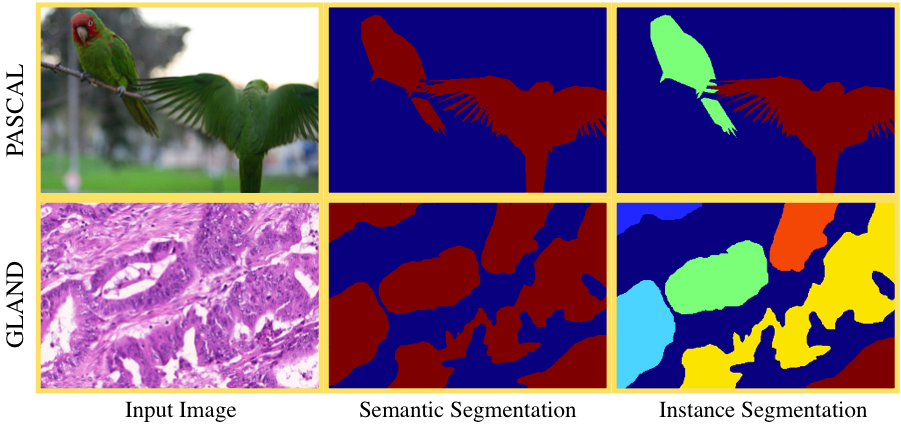} 
\end{tabular}
\caption{\footnotesize Illustration of the instance segmentation problem we are tackling here. The first, the second, and the third column show input images, the semantic labeling maps, and the corresponding instance labeling maps, respectively. The first and the second row display typical examples from PASCAL VOC 2012 \cite{pascal-voc-2012}, and MICCAI 2015 Gland segmentation dataset \cite{miccai2015gland}, respectively.}
\label{fig:overview}
\end{center}
\vspace{-7mm}
\end{figure}

Instance segmentation has recently become an important task, which is more challenging than both object detection and semantic segmentation tasks, since its goal is to label as well as provide pixel-level segmentation to each object in the image. 
Figure \ref{fig:overview} shows an illustration of intance segmentation on images from two benchmark datasets, namely PASCAL VOC \cite{pascal-voc-2012}, and the gland segmentation benchmark \cite{miccai2015gland}.
In the standard semantic labeling task \cite{shotton2006textonboost}, each pixel in an image is assigned with an object class label e.g. sky, road, car etc.; in the instance segmentation problem, each pixel is additionally associated with an instance ID indicating which object it belongs to. Therefore, there are two sets of labeling maps for each image: (1) a semantic class labeling map (this is a classification problem), and (2) an instance ID labeling map (this is a segmentation problem).
%Object class labeling can be conveniently formulated as a classification problem where methods like fully convolutional neural networks (FCN) \cite{long2015fully} can be applied to performed pixel-wise prediction. Instance labeling however is difficult: the exact ID assigned to each instance is not important and unique; there exists a quotient space and for example car$\#1$ and car$\#2$ can respectively be named as car$\#2$ and car$\#1$ instead; instance labeling is a segmentation problem and the goal here is to separate all the object instances in an image, which prohibits us from training a direct classifier for the instance IDs.

For the remainder of this paper, we refer to {\em semantic labeling} as the task of predicting per-pixel object class label and refer to {\em instance labeling} as the job of assigning an instance ID to each region. On one hand, object labeling and instance labeling are two different tasks, as explained above. On the other hand, the two tasks are highly correlated. If every object instance forms a connected component (not being cut into two disjoint parts due to occlusion which rarely happens in practice) and no two objects belonging to the same class are connected to each other, then a semantic labeling map can be readily converted into an instance labeling map by obtaining the connected component of each instance. This happens frequently but it is not always true, as we can see in Figure \ref{fig:overview}.
% which is the reason why FCN can be used as a baseline for instance segmentation. %Overall, directly formulating instance segmentation into multi-task classification is not ideal too.

Instance segmentation methods can be roughly divided into two categories: (1) those detection-based methods that perform bounding box detection \cite{pinheiro2015learning,SharpMask,dai2015instance, dai2016instance,hariharan2015hypercolumn, liang2015proposal,li2016instance}; and (2) those segmentation-based methods that use dense per-pixel features \cite{silberman2014instance,riemenschneider2012hough,zhang2015instance}. Detection-based methods typically perform proposal and object detection, followed by instance masking. These methods require objects being tightly bounded by rectangular boxes, which is a strong condition to satisfy. 
Existing segmentation-based methods avoid the object detection stage but their application domains are not as general as the PASCAL VOC scenarios, e.g. only one foreground object class in \cite{riemenschneider2012hough,zhang2015instance} or using the NYU dataset where additional depth information is available \cite{silberman2014instance}.
%Most existing frameworks for instance segmentation is based on detection, such as Hypercolumn \cite{hariharan2015hypercolumn}, MNC \cite{dai2015instance,dai2016instance}, DeepMask \cite{pinheiro2015learning}, which requires to generate thousands of proposals for further detection. PFN \cite{liang2015proposal} is a proposal-free network, however it still needs to regress the bounding box location of the instance objects. MPA \cite{MPA2016} aggreates mid-level patch segment prediction results. There are also instance segmentation for other applications, such as indoor scenes \cite{silberman2014instance} and autonomous driving \cite{zhang2015instance}. 
%to be of large practical importance as image labeling cannot separate two adjacent objects that share the same label.
%The connection between semantic labeling and instance labeling has been mostly overlooked and recent methods developed in this domain primarily use a

%Recent efforts adopting FCN for instance segmentation \cite{liang2015proposal,dai2016instance} are still fundamentally detection-based that do not study how pixels on different instances are grouped or segmented.

In this paper, we study the fundamental challenge in instance segmentation and aim to develop object proposal- and detection- free methods. The reason for us to avoid the object detection process is twofold: (1) predicting object bounding boxes \cite{girshick2014rich} and labeling individual pixels \cite{long2015fully} are two different tasks that involve fairly different modules, which results in large complexity in both training and testing when combining the two; (2) although fast object detectors \cite{ren2015faster} are being invented but they are fundamentally limited in making dense pixel-level labeling. Existing instance segmentation methods, however, follow a common thread by performing object detection first with additional segmentation masking \cite{dai2015instance,SharpMask}.

The fully convolutional neural networks (FCN) family models \cite{long2015fully,CP2015Semantic,CP2016Deeplab,yu2015multi} have shown the significant benefit in making dense prediction, allowing end-to-end learning for pixel-level classification and regression. The problem of image instance segmentation, however, cannot be directly formulated as a classification problem. In instance segmentation, the region ID has no direct semantic meaning and the exact ID assigned to each region is not important and unique: there exists a quotient space for the labeling; e.g. car $\#1$ and car $\#2$ can respectively be named as car $\#2$ and car $\#1$ instead. So, we need to make transformations of instance labeling to a formulation that can be tackled by as a classification/regression problem.
%further formulate the problem, and then the powerful convolutional neural networks \cite{lecun1989backpropagation} can be applied in learning these complex patterns.

One possibility to leverage the power of convolutional neural networks \cite{lecun1989backpropagation} in learning complex patterns is to formulate the segmentation problem as affinity learning \cite{fowlkes2003learning}: pixels belonging to the same segment receive a high affinity score (e.g. $1$) and those from different segments have a low affinity score (e.g. $0$). Moreover, the affinity learning problem can be in pixel-level or superpixel-level. CNN-family models have been successfully used to learn face similarity \cite{taigman2014deepface,sun2014deep}; FCN-like models have been used to find correspondences for the flow fields when matching two images\cite{fischer2015flownet}. However, learning pixel affinity within the same image is not an easy task in practice, considering the large number of pixel pairs, and some special care needs to be taken even for the foreground/background segregation problem \cite{maire2015affinity}.

%One way to transform the instance labeling is to formulate the segmentation problem as affinity learning \cite{fowlkes2003learning}: pixels belonging to the same segment receive a high affinity score (e.g. $1$) and those from different segments have a low affinity score (e.g. $0$). Moreover, the affinity learning problem can be in pixel-level or superpixel-level. CNN-family models have been successfully used to learn face similarity \cite{taigman2014deepface,sun2014deep}; FCN-like models have been used to find correspondences for the flow fields when matching two images\cite{fischer2015flownet}. However, learning pixel affinity within the same image is not an easy task in practice, considering the large number of pixel pairs, and some special care needs to be taken even for the foreground/background segregation problem \cite{maire2015affinity}.

% ZT: this paragraph is confusing if stated here
%Another perspective to transform the instance labeling is to formulate the segmentation problem as instance boundary detection: pixels that lie on the instance boundaries are labeled as edge (e.g. $1$) and those inside the instances or in the background have labeled as non-edge (e.g. $0$). CNN-family models \cite{xie2015holistically} have been successfully used in edge detection problems for natural images. Here, we apply these models for instance boundary detection in order to generate instance segmentation.

\begin{figure*}[!htp]
\begin{center}
\begin{tabular} {c}
\includegraphics[width=0.95\linewidth]{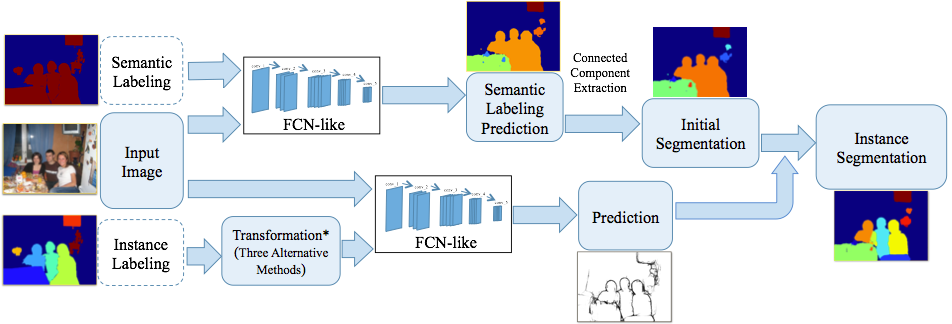} 
\end{tabular}
\caption{\footnotesize Illustration of the pipeline of our overall method. Each input image is associated with a semantic labeling map and an instance labeling map, as shown in Figure \ref{fig:overview}. These two labeling maps are only presented during training. Our method has two basic paths that are both built on fully convolutional neural networks. The first path learns a FCN-like model for the pixel-wise classification by taking the input image and its semantic labeling; the semantic labeling prediction can be used to create an initial segmentation using a connected component extraction procedure. In the second path, we provide three alternative methods to transform the instance labeling; three alternative methods are (1) pixel-based affinity mapping, (2) superpixel-based affinity learning, and (3) boundary-based component segmentation. Another FCN-like model is trained on the input image and the transformed instance labeling. The prediction from instance labeling transformation path is further integrated with the initial segmentation extracted from semantic labeling path to generate the final instance segmentation results. Some typical examples of the results are shown in Figure \ref{fig:gland_results}.}
\label{fig:pipeline}
\end{center}
\vspace{-6mm}
\end{figure*}

\begin{figure*}[!htp]
\begin{center}
\begin{tabular} {c}
\includegraphics[width=1.0\linewidth]{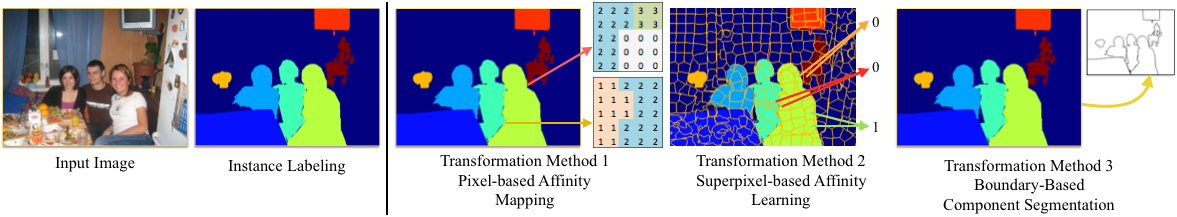} 
\end{tabular}
\caption{\footnotesize Illustration of our proposed three alternative instance labeling transformation methods. The first column is the input image, and the second column is its instance labeling ground truth. Our proposed three alternative instance labeling transformation methods are shown in the third, fourth, fifth column, respectively. Transformation method 1 (pixel-based affinity mapping) will map the pixel-based local affinity pattern to a specific class (Section \ref{sec:pixel}). Transformation method 2 (superpixel-based affinity learning) will generate affinities of superpixel pairs (Section \ref{sec:sp}). Transformation method 3 (boundary-based component segmentation) will produce the instance boundaries (Section \ref{sec:edge}).}
\label{fig:transformations}
\end{center}
\vspace{-7mm}
\end{figure*}

In this paper, we propose an object proposal- and detection- free segmentation-based framework for instance segmentation problem. Our framework consists of two paths, as shown in Figure \ref{fig:pipeline}. The semantic labeling path focuses on the pixel classification problem. To handle the instance labeling quotient space, we introduce the second path for instance labeling transformation and prediction. Here, we explore and develop three new methods for instance labeling transformation (as shown in Figure \ref{fig:transformations}): (1) {\em pixel-based affinity mapping}, (2) {\em superpixel-based affinity learning}, and (3) {\em boundary-based component segmentation}. The predictions from the instance labeling transformation path are integrated with the semantic labeling path to generate instance segmentation results.
% we explore and develop three new strategy alternatives

Our framework is object proposal- and detection-free, which is simpler and more transparent than existing frameworks. Also, three new instance labeling transformation methods have been proposed. 
Using a similar network structure, we are able to produce competitive results on two different types of datasets, namely PASCAL VOC \cite{pascal-voc-2012} which is object-centric, and the gland segmentation benchmark \cite{miccai2015gland} which is texture-centric. We achieve competitive results on both datasets, which the existing methods fail to do.
For example, due to the assumption about tightly bounded single object within a rectangle,  detection-based method is not able to achieve good performance (see the gland instance segmentation task example in Figure \ref{fig:overview} and experimental results in Section \ref{sec:gland}).

\noindent
\textbf{The significance of being object proposal- and detection- free}

%In this paper, we develop a object proposal- and detection- free algorithm for instance segmentation by addressing a fundamental challenge of transformation instance labeling.
% by addressing a fundamental challenge of learning pair-wise pixel affinity for instance segmentation.
% using CNN, or more specifically FCN-like methods \cite{long2015fully,xie2015holistically,yu2015multi}.
Our motivation to develop an object proposal- and detection- free approach for instance segmentation is threefold.
First, detection-based methods require objects to be non-deforming and capable of being tightly bounded by rectangular bounding boxes. These are strong preconditions that produce great limitations. For example, adopting the system \cite{dai2015instance} to perform gland instance segmentation shows unfavorable results (see results in Section \ref{sec:gland}), since the glands have large deformation that are hard to be bounded by rectangular boxes.
Second, detection-based methods involve many additional steps, making the algorithm complex and opaque. Third, proposal- and detection- based methods do not perform direct segmentation and have fundamental limitations when multiple objects interact and appear in the same bounding box.

%These methods require objects being tightly bounded by rectangular bounding boxes. This is a strong condition to satisfy and it potentially limits detection-based methods to be further extended to dealing with a large variety of objects in generic scenes. 

%-------------------------------------------------------------------------
\section{Related Work}

\noindent
\textbf{Semantic Segmentation} Deep convolutional neural networks advances the progress of semantic segmentation research. Some recent models focus on using fully convolutional networks for the dense pixel-wise prediction \cite{long2015fully,CP2015Semantic,CP2016Deeplab,yu2015multi}. `Atrous convolution' \cite{CP2015Semantic,yu2015multi} has been proved to be effective for explicitly enlarge the receptive field. CRF models can be applied in post-processing \cite{CP2015Semantic} or in-network \cite{zheng2015conditional} to refine segmentation contours. Semantic segmentation is part of our proposed framework, so the advances from these segmentation models will benefit our instance segmentation approach as well.

\noindent
\textbf{Affinity Learning} Spectral clustering methods, such as normalized cuts (Ncuts)~\cite{shi2000normalized}, have shown to be effective for unsupervised segmentation. However, computing accurate affinity matrix is a keystone but also a handicap for spectral clustering algorithms. In the past, the affinity matrix is mostly calculated based on hand-designed heuristics~\cite{shi2000normalized}. Previous attempts in learning the affinities \cite{fowlkes2003learning}, while being inspiring, have not shown to significantly benefit segmentation. CNN-based approach has been used for foreground and background segregation \cite{maire2015affinity}. Our method focuses on the affinity mapping in order to make transformation of instance labeling.

\noindent
\textbf{Instance Segmentation} 
Methods for instance segmentation can be roughly divided into two categories: (1) detection-based methods that perform bounding box detection \cite{pinheiro2015learning,SharpMask,dai2015instance, dai2016instance,hariharan2015hypercolumn, liang2015proposal,li2016instance}; and (2) segmentation-based methods that use dense per-pixel features \cite{silberman2014instance,riemenschneider2012hough,zhang2015instance}.
%Most existing frameworks for instance segmentation is based on detection, such as Hypercolumn \cite{hariharan2015hypercolumn}, MNC \cite{dai2015instance,dai2016instance}, DeepMask \cite{pinheiro2015learning}, which requires to generate thousands of proposals for further detection. PFN \cite{liang2015proposal} is a proposal-free network, however it still needs to regress the bounding box location of the instance objects. MPA \cite{MPA2016} aggreates mid-level patch segment prediction results. There are also instance segmentation for other applications, such as indoor scenes \cite{silberman2014instance} and autonomous driving \cite{zhang2015instance}. 
%to be of large practical importance as image labeling cannot separate two adjacent objects that share the same label.
%The connection between semantic labeling and instance labeling has been mostly overlooked and recent methods developed in this domain primarily use a

Detection-based methods typically perform proposal and object detection, followed by instance masking.  
The Deep Mask method \cite{pinheiro2015learning} generates object proposals, followed by learning a binary mask for each detected bounding box; a cascade strategy is adopted in \cite{dai2015instance} for instance localization and then masking. 
PFN \cite{liang2015proposal} is a proposal-free network, however it needs to regress the bounding box locations of the instance objects. MPA \cite{MPA2016} aggregates mid-level patch segment prediction results by sliding on the feature maps. 

Existing segmentation-based methods avoid the object detection stage but their application domains are not as general as the PASCAL VOC scenarios, e.g. assuming one foreground object class only in \cite{riemenschneider2012hough,zhang2015instance} or using the NYU dataset where additional depth information is available \cite{silberman2014instance}.
%Recent efforts adopting FCN for instance segmentation \cite{liang2015proposal,dai2016instance} are still fundamentally detection-based that do not study how pixels on different instances are grouped or segmented.
In \cite{riemenschneider2012hough}, a Hough space is created to perform the segmentation; in \cite{silberman2014instance} a structured labeling formulated is proposed to explore a segmentation tree; in \cite{zhang2015instance} regions/instances are assigned with depth order, allowing classification to be learned.

Our framework is object proposal- and detection-free, which is more simple and transparent than existing frameworks. Also, our approach focuses on the transformations of instance labeling and proposes three different methods. 
% Also, we will show it works on both natural images, such as PASCAL VOC as well as gland instance images. Since the glands observe large deformation that are hard to be bounded in rectangular boxes, most detection-based methods achieve unfavorable results.

%-------------------------------------------------------------------------
\section{Instance Segmentation Framework}
In this section, we discuss our proposed framework for object proposal-free and object detection-free  instance segmentation. Our framework consists of two fully convolutional neural network paths, as shown in Figure \ref{fig:pipeline}. The semantic labeling path focuses on the per-pixel classification problem, which predicts the category label for each pixel. To tackle the instance labeling quotient space problem, we introduce the other path for instance labeling transformation and prediction. Our system takes the original images as input and trains two FCN modules to predict the semantic labels and the instance label transformation maps separately. The predictions from instance labeling transformation path are then integrated with the semantic labeling path to generate instance segmentation results.

\subsection{Semantic Labeling}
The main goal of semantic labeling is to predict a detailed mask in pixel-level for different classes. FCN-based semantic segmentation models \cite{long2015fully,CP2016Deeplab,yu2015multi} have achieved rapid process in recent years. In our framework, this semantic labeling path takes the input image and its semantic labeling as training data and performs per-pixel prediction. We adopt the `DeepLab-Large-FOV' network structure \cite{CP2016Deeplab} as the basic network for our semantic labeling path since it delivers state-of-the-art performance. This FCN network introduces `atrous convolution', which can explicitly control the resolution of the feature responses and effectively enlarge the field of view of filters to incorporate larger context. As we don't observe much difference in the performance using the dense CRF post-processing, so we drop the CRF processing step. 
%and out main attention is focused on the instance labeling transformation 
% Our semantic labeling network is a modified version of DeepLab network structure \cite{CP2015Semantic,CP2016Deeplab}. 

%Semantic segmentation, especially FCN-like models \cite{long2015fully,CP2016Deeplab}, has achieved rapid progress in recent years. 
%To perform semantic labeling, FCN-like models \cite{long2015fully,yu2015multi} are able to deliver state-of-the-art results; our main attention of this paper is therefore given to describing LLAM for instance segmentation.
%Our main motivation is to readily apply a detection-free method like fully convolutional networks method \cite{long2015fully} to perform dense pixel-wise regression by embedding a neighborhood labeling map into a one-dimensional space.  %Our system mainly consists of these components: label affinity mapping, affinity map prediction, semantic labeling prediction, initial instance segmentation extraction and the final combination of the instance segmentation.

\subsection{Instance Labeling Transformation}

Swapping the instance IDs will lead to the same instance results, which causes the quotient space problem (see Section \ref{sec:instance} for more details). So we need to transform the instance labeling to further formulate the problem. We propose three transformation methods (as shown in Figure \ref{fig:transformations}), which are all segmentation-based and object proposal- and detection-free. We find that these methods are effective for instance labeling transformation. 

Affinity is the natural choice to measure the coherence of pixels in the images. From the perspective of metric learning, pixels in the same region should have small distances, hence large affinities/similarities, and pixels in different regions should have large distances, hence small affinities/similarities. Using CNN classification offers a possible solution to learn the affinity patterns in principle. We propose two affinity based transformation methods, which are both clustering based methods. The first one is {\em pixel-based affinity mapping} and the other is {\em super-pixel affinity learning}, which will be discussed in more details in Section \ref{sec:pixel} and \ref{sec:sp}, respectively.

Object boundaries provide another perspective for instance labeling transformation as they provide the cues to locate the object instances. So, our third strategy is {\em boundary-based component segmentation}, which is a non-clustering method. Its idea is to leverage the instance boundary to separate different instances in the same component from semantic labeling prediction. We will discuss this method in Section \ref{sec:edge}.

\subsection{Integrate Instance and Semantic Labeling}

Our initial segmentation results are from the semantic labeling. We extract the connected components of the same category and regard them as the potential instances. Then, we can utilize our learned instance labeling prediction to separate the neighbor instances as shown in Figure \ref{fig:overview}. We notice that in many cases, these connected components provide a good starting point for the instance segmentation task. So we start from the segmentation perspective, which is different from object proposal-and-detection based methods. Different instance labeling transformation methods will have different ways to integrate with semantic labeling predictions, which will also be discussed in Section \ref{sec:instance}.
%refine the instance segmentation results,

%So it comes with several benefits. The first one is the simplicity and transparency of our framework. Proposal-based instance segmentation methods requires to generate thousands of proposals and involve many additional steps, making the algorithm complex and opaque. 

%-------------------------------------------------------------------------
\section{Instance Labeling Transformations \label{sec:instance}}

%In instance segmentation, each object instance is given a unique ID to identify the specific instance. However, the exact instance ID values are not that important. For example, randomly swapping IDs in the same image will lead to the same instance segmentation result, so there exists a quotient space for labeling. So, CNN-based models can not be directly used to predict the instance labeling and transformation of instance labeling is needed. In this section, we propose three options to make transformations of instance labeling and adopt powerful CNN models to learn these transformations. Moreover, we discuss how to integrate these transformations with semantic labeling to generate instance segmentation results.

For the instance segmentation problem, we are given a training set $S=\{(X_n, Y_n, A_n); n=1...N\}$ where $X_n$ refers to the $n$-th input image, $Y_n$ refers to its corresponding semantic labeling, $A_n$ refers to its corresponding instance labeling, and $N$ denotes the total number of images in the training set. For simplicity of annotation, we use $(X, Y, A)$ by subsequently dropping the index to focus on one input.  Grouping pixels with the same instance label allows us to have another representation, regions denoted by $\mathcal{R} = \{R_m; m=1...M\}$, where $M$ refers to the total number of regions in $\mathcal{R}$, $R_{m_1} \cap R_{m_2} = \emptyset, \forall m_1 \ne m_2$, and $\cup_{m=1..M} R_m = \Omega$. $\Omega$ includes all the pixels in the image. It is worth mentioning that $A$ and $\mathcal{R}$ is a many-to-one mapping with a quotient space, which is exactly one source of the challenges in instance segmentation being tackled here. We explain in detail below.
For example, if we assign labels to $\mathcal{R}$ as $(1,2,3,...,M)$ to $(R_1, R_2,...,R_M)$ respectively, then we obtain one instance labeling for image $X$ denoted as $A^{(1)}=(a_i=k \;if\; i\in R_k;i=1...|X|)$ where $i$ indexes each pixel $i$, $a_i$ is the instance label of pixel $i$, and $|X|$ is the total number of pixels of $X$. However, with the same $\mathcal{R}$, we instead assign as $(2,1,3,...,M)$ to $(R_1, R_2,...,R_M)$, then the instance labeling would be $A^{(2)}=(a_i=1 \;if\; i\in R_2, \;and\; a_i=2 \;if\; i\in R_1, \;else\; a_i = k \;if\; i \in R_k; i=1...|X|)$. Therefore, we can see both $A^{(1)}$ and $A^{(2)}$ refer to the same $\mathcal{R}$. For this reason, we propose instance labeling transformation methods that map all different $A$s correspondingly to the same/similar $\mathcal{R}$ into a new form, which can be tackled by a classification/regression algorithm (here a FCN-like model).

%Each region $R_m$ has its corresponding instance label $v_m$. So we can have the mapping from the regions to their instance labels as $\mathbf{A}(R_m) = v_m$.
%Also, for any two regions $R_i$ and $R_j$ ($i \neq j$), the intersection of $R_i$ and $R_j$ is $\emptyset$. 
%By swapping the instance labels $v_m$, we can get new instance labels for each region and have the mapping $\mathbf{B}$, which is different from $\mathbf{A}$. And different swapping process will lead to different instance label mappings.  So the exact instance IDs are not important and there exists a quotient space for instance labeling.

%Therefore, CNN-based models can not be directly used to predict the instance labeling and transformations of instance labeling is in need.
Next, we preset three alternative methods to transform instance labeling (as shown in Figure \ref{fig:transformations}) and apply FCN-based models to make predictions.

\subsection{Method 1: pixel-based affinity mapping \label{sec:pixel}}

The first option is to perform clustering/segmentation based on pair-wise pixel affinity to tackle the quotient space problem.
%One possible way of grouping similar pixels is to obtain their pair-wise similarity, which is also known as affinity.
Given the ground truth of an instance labeling map, we can construct the global affinity matrix. However, learning and computing the global affinity is both computationally expensive and practically infeasible. For this reason, some special care is taken in \cite{maire2015affinity} for foreground and background segregation.

Here, we focus on local affinity patterns and develop a novel affinity learning method by transforming the instance labeling map into different classes. Thus, an instance labeling map is turned into a classification map, in which each pixel is associated with a class, indicating a local affinity pattern. In this way, we are able to train fully convolutional networks to perform pixel-based classification to obtain a comprehensive affinity map for the entire image.

\begin{figure}[!htp]
\begin{center}
\begin{tabular} {c}
\includegraphics[width=1.0\linewidth]{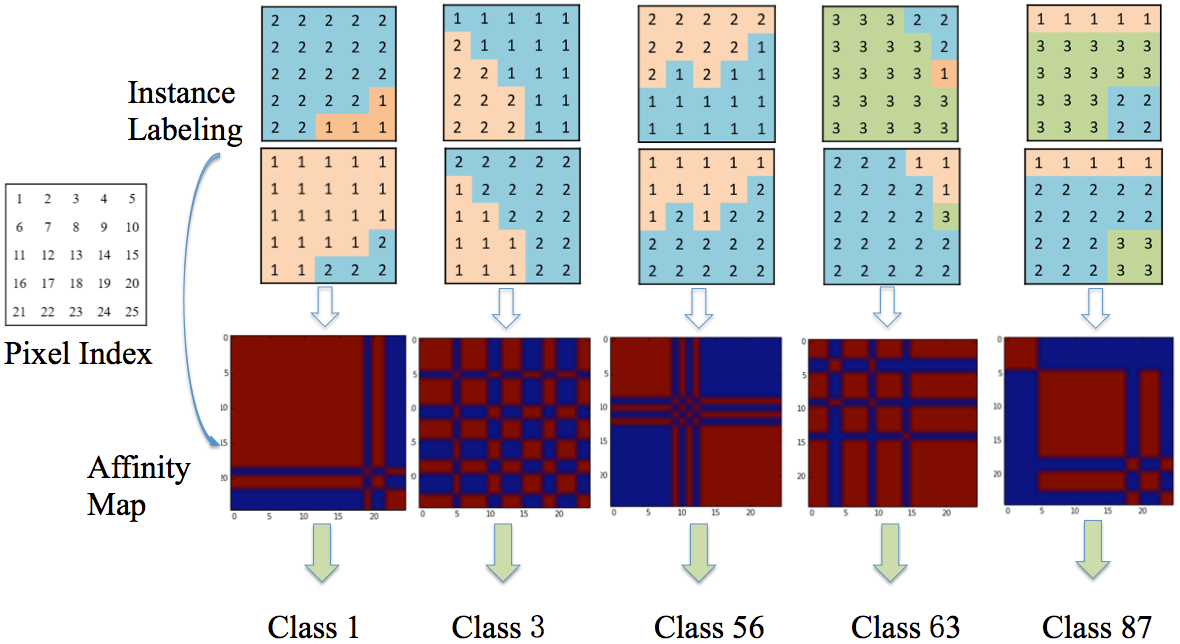} 
\end{tabular}
\caption{\footnotesize Illustration of performing pixel-based affinity mapping. We use $k=5$ in the paper. Given an instance labeling map, every image patch of size $5 \times 5$ is first extracted (this is shown in the first row); each $5 \times 5$ labeling configuration has an associated affinity map defined on the labels for the total $25$ pixels in the patch, resulting in an affinity matrix of size $25 \times 25$ (this is shown in the third row); the pixel index for the affinity matrix is shown on the left. Based on the element-wise distance of the resulting affinity matrices, they are clustered into 100 classes by $k$-means. The first and the second row show that two different instance labeling for the same segmentation will lead to the same affinity map, and hence the same class, which is desirable and it makes learning a classification model feasible.}
\label{fig:nlam}
\end{center}
\end{figure}

\noindent
{\bf Local affinity mapping}
Consider the possible patterns that appear in a $k\times k$ image patch, which centers at pixel $(i,j)$. Here, $k$ is the size of the patch and $(i, j)$ are the coordinates of the center pixel of the $k\times k$ patch. Given that there are $n$ different instances that appear within the same patch, there are a combinatorial number, $n^{k\times k}$ possible cases. This leads to a large number, even if $k$ and $n$ are relatively small. However, in the presence of the quotient space of the labeling maps, swapping the instance IDs of any two regions gives rise to the identical segmentation result. As Figure~\ref{fig:nlam} shows, different configurations of instance labeling (in row 1 and row 2) will lead to the same affinity pattern (in row 3). Therefore, we can leverage this property to simplify the problem at hand.
%it is not necessary to compute every possible configuration, but to find instance configurations that appear most frequently in the images. 
Here, each $k \times k$ labeling configuration is firstly associated with a local affinity map, which is defined on the labels for every pair of the total $k^2$ pixels in the patch. Therefore, the size of the local affinity matrix is $k^2 \times k^2$. Next, we adopt $k$-means to project the high dimensional local affinity matrix into a fixed number of classes. This embedding assigns each local labeling pattern with a class.

\noindent
{\bf Training and prediction} 
To train our model, we construct the pixel-based affinity mapping from the ground-truth instance labeling. In our experiments, we firstly resize the original images to a smaller size and then calculate the pixel-based affinity mapping in a fixed-size patch. We select the scale of the resized images as $64 \times 64$ and adopt $5 \times 5$ patch. Here, we consider the pair-wise relationship between each pixel in this patch, so we can construct the local affinity matrix with the size of $25 \times 25$  in our setting. Then, we adopt $k$-means to identify the classes of these affinity patterns. This mapping process is shown in Figure \ref{fig:nlam}. After this mapping, each pixel is given a new affinity label and its value indicates the local patterns. In this way, we formulate this problem to a pixel-based classification task, which is suitable for FCN to solve. We modify the FCN network architecture \cite{long2015fully} by removing the deconvolution layer in FCN and modifying the stride of the last two pooling layers to 1. The cross entropy loss is used to train the network. 
% to construct local affinity maps
% so we can construct the local affinity map as the size of 625

% From our experiments, we demonstrate that this simple network achieves promising results.

%We modify the basic FCN-like architecture~\cite{long2015fully} to predict the affinity map. The input of our network is fixed to $480 \times 480$ and the output affinity label map of the network is $64 \times 64$. So, we remove the deconvolution layer in FCN \cite{long2015fully} and modify the stride of the last two pooling layers to 1. It has the 5-stage structure and we use the weights of the pre-trained VGG network to initialize the network. The cross entropy loss is used to train the network. From our experiments, we demonstrate that this simple network achieves promising results.
%As we are doing the classification task, cross entropy loss is used to train the network. From our experiments, we demonstrate that this simple network achieves promising results.

In the next step, we can use the learned pixel-based affinity values to reconstruct the local affinity matrix. Each class label corresponds to the local affinity matrix. Then, we need to use these pixel-based affinity labels to fill in the overall affinity matrix. In our setting, each pixel-based affinity label will affect the affinity relationship among all the pixels in the same patch. As two neighbor pixels can appear in different patches, their pairwise pixel affinity is voted by all these local information gathered from these different patches. Finally, we construct the overall local affinity matrix, which will be used for distinguishing different instances.
% existing high-dimensional affinity vector, which appears in the images. 

\noindent
{\bf Integration with semantic labeling}
For each single connected component in each category, we firstly extract the affinities of the pixels belonging to that component, and then apply spectral clustering algorithms, such as the normalized cut (NCuts) algorithm \cite{shi2000normalized}, to the local affinity matrices. By varying the number of cuts, we can obtain multiple potential instance settings for that component. Finally, we put the local instance settings back to the global instance labeling map. After we go through all the single connected components of all the categories, the refined instance segmentation results are obtained. 

\subsection{Method 2: superpixel-based affinity learning \label{sec:sp}}

As we discussed in Section \ref{sec:pixel}, computing the global affinity matrix is not feasible in practice, in face of the huge number of pixel pairs. Our second transformation method is to directly measure the affinity between superpixels. Leveraging superpixels have brought some benefits. First, superpixels are more natural and meaningful representation of visual scenes, which simplifies the low-level pixel grouping process. Second, it can reduce the complexity of affinity computation. 

After generating superpixels from the original images, we can assign the affinity for superpixel pairs. Specifically, for super-pixels with the same instance labels, we assign the affinity metric as 1, and for super-pixels with different instance labels, we assign the affinity metric as 0. This process is shown in the fourth image of Figure \ref{fig:transformations}. So, for an input image, we can get the affinity labels of all the superpixel pairs as a vector. In this way, we formulate the problem which can be learned by FCN-like models.

\noindent
{\bf Training and prediction}
In our experiment, we adopt the SLIC method \cite{SLIC} to generate superpixels and then construct the affinity for these superpixel pairs. We modify the FCN network architecture~\cite{long2015fully} to learn the superpixel affinity. For each superpixel, we randomly select a fixed number of pixels inside it to calculate the average feature map from the FCNs. Then, for a pair of superpixels, we concatenate their feature maps, and pass them to two newly added convolution layers to make the affinity prediction. The cross entropy loss is used to train the network. We make this network to be trained in an end-to-end way. The learned superpixel affinity metric can be used to construct the local affinity matrix directly, without the voting process as pixel-based affinity mapping has.

% We modify the 16-layer VGG architecture \cite{simonyan2015very} to predict the affinity metric. 
% As we are doing the classification task,

\noindent
{\bf Integration with semantic labeling}
The integration process is similar to pixel-based affinity mapping method. For each single connected component of each category, we firstly extract the superpixels as well as the affinity predictions of these superpixel pairs. Then, we can construct the local affinity matrices, and spectral clustering algorithms, such as the normalized cut (NCuts) algorithm \cite{shi2000normalized}, are used to obtain potential instances. Then, we put the local instance settings back to the global instance labeling map. After we go through all the single connected components of all the categories, the final instance segmentation results for the image are generated. 

\subsection{Method 3: boundary-based component segmentation \label{sec:edge}}

Another observation of the quotient space is that no matter how we swap the instance IDs, the instance boundaries remain the same. This intrinsic property provides us with the possibility to transform the instance labeling into instance boundaries. By this transformation, edge detection methods can be applied to learn these specific instance boundaries. After obtaining predicted instance boundaries, we can use these edges to separate connected component from semantic labeling. The benefit of boundary-based component segmentation is that we don't need to identify the number of instances as clustering based methods do. This edge-based componenet segmentation method is very simple and achieves results on par with state-of-the-art performance on the gland segmentation dataset.

% The instance boundary is where instance IDs change.

\noindent
{\bf Training and prediction}
We first generate the instance boundary labels from the ground truth instance labeling, as shown in Figure \ref{fig:transformations}. Then, we adopt the recent FCN-based boundary detection model, such as HED \cite{xie2015holistically}, to learn the instance boundary. HED provides the holistic network to learn multi-scale and multi-level features for boundary detection. It combines fully convolutional neural networks \cite{long2015fully} and deeply-supervised nets \cite{lee2015deeply} to perform image-to-image prediction. For the boundary maps computed from HED, we also apply a standard non-maximal suppression technique to obtain thinned boundaries. Though HED is designed for boundaries in natural images, from our experiments, we show this network structure is also very effective for locating specific instance boundaries.

\noindent
{\bf Integration with semantic labeling}
We adopt a simple method to integration the results from the predicted instance boundaries and semantic labeling. For the pixels that are predicted as boundaries, we simply assign their corresponding pixels in the semantic labeling as background label. Thus, these predicted boundaries can separate instances in the same connected component of semantic labeling if there exits a complete boundary inside the component. Finally, we get all the connected component in the updated semantic labeling map to generate the instance segmentation results.

%---------------------------------------------------------------------------

\section{Experimental Results}

%\subsection{Implementation}
In this section we evaluate the performance of our proposed approach on two instance segmentation benchmarks, PASCAL VOC 2012 \cite{pascal-voc-2012} and MICCAI 2015 gland segmentation dataset. These two datasets are object-centric and texture-centric, respectively. For training and fine-tuning our network, our system is built upon the Caffe platform \cite{jia2014caffe}. We use the released VGG-16 \cite{simonyan2015very} model to initialize the convolution layers in our network. While for other new layers, we randomly initialize them by sampling from a zero-mean Gaussian distribution. Our training complexity is similar to those reported FCN-like models. Our testing speed, in particular method 2 and method 3, is very fast (less than 1 second).

%The initial learning rate is 0.001 and the batch size is set to 10. 
%The momentum value is 0.9 and weight decay ratio is 0.001.

\subsection{PASCAL VOC 2012}

We evaluate our approach on PASCAL VOC 2012 dataset \cite{pascal-voc-2012}. We use the augmented dataset from SBD\cite{BharathICCV2011} and collect the instance labels from \cite{pascal-voc-2012} and \cite{BharathICCV2011} for training. At the test stage, we measure our performance on PASCAL VOC 2012 segmentation validation set. Two standard evalution metrics, $AP^r$ and AR@N, are used for comparison.

The first evaluation metric is $AP^r$, which measures the average precision under 0.5 IoU overlap with ground-truth segmentation. The evaluation results is summarized in Table \ref{tb:voc_map}. Here, FrontEnd (FE) refers to extracting connected components directly from semantic labeling predictions. The boundary-based component segmentation (Method 3) achieves the best performance in our three transformation methods. Also, all three transformation methods outperform detection-based models, SDS \cite{hariharan2014simultaneous} and the method in \cite{Chen_2015_CVPR}. However, they are worse than recent models, PFN \cite{liang2015proposal}, MPA \cite{MPA2016} and R2-IOS \cite{Liang2016}. One reason is that it is hard to give an accurate number of instances for clustering based methods (method 1 and method 2). Also, we simply use the area of each instance as the score for evaluation, which can not be optimal. Our framework is object proposal-free and detection-free, and much simpler than their models, which have multiple components. 

\newcolumntype{C}[1]{>{\centering\let\newline\\\arraybackslash\hspace{0pt}}m{#1}}
\begin{table}[!htp] \small
\begin{center}
\begin{tabular} {c|c}
Method & $AP^r$ ($\%$)\\
\hline
{\bf Proposal-and-detection based} \\
SDS \cite{hariharan2014simultaneous} & $43.9$ \\
Chen et al. \cite{Chen_2015_CVPR}  & $46.3$ \\
R2-IOS \cite{Liang2016} & $66.7$ \\
\hline
\hline
PFN unified \cite{liang2015proposal} (w/ object localization) & $49.1$\\
PFN independent \cite{liang2015proposal} (w/ object localization) & $58.7$ \\
MPA 3-scale \cite{MPA2016} (w/ sliding window) & $62.1$ \\
\hline
\hline
{\bf Segmentation based} \\
Direct labeling (FE \cite{CP2015Semantic}) & $45.3$ \\
\hline
FE+Method 1 (ours) & $46.7$  \\
FE+Method 2 (ours) & $48.1$  \\
FE+Method 3 (ours) & $49.9$ 
\end{tabular}
\caption{\footnotesize Comparison on the PASCAL VOC instance segmentation validation set based on $AP^r$. Models are grouped into proposal-and-detection based and segmentation based. FrontEnd (FE) refers to extracting connected components directly from semantic labeling prediction. For the transformation methods, Method 1 refers to pixel-based affinity mapping, Method 2 refers to superpixel-based affinity learning, and Method 3 refers to boundary-based component segmentation. Our framework is object proposal free and detection free, and much simpler than other methods, which have multiple components. Note that PFN uses object localization and MPA has a sliding window procedure, which are both object detection like.}
\label{tb:voc_map}
\end{center}
\end{table}
% Models are grouped into proposal-and-detection based, object localization based, sliding window based and segmentation based.

% Our simple instance segmentation framework achieves promising results.

%In this experiment, we compare our three strategy with SDS \cite{hariharan2014simultaneous}, Chen et al. \cite{}

The second evaluation metric is AR@N, which measures the average recall between IoU overlap threshold from 0.5 to 1.0. The evaluation results is summarized in Table \ref{tb:voc12_ar}. Our proposed framework achieves the comparable results with proposal-based methods, especially in AR@10. Since our framework is object proposal- and detection- free, it doesn't generate hundreds of proposals, leading to a relatively lower score on AR@100.
% Our result with AR@10 measure that is fully segmentation based is comparable with other approaches that engage sliding windows for object detection.
% how to give k is a hard problem

%For the semantic labeling, we adopt the recent model~\cite{yu2015multi} which is pre-trained on PASCAL VOC 2012. Additionally, we trained a separate FCN for predicting the affinity map. 

%Firstly, we adopt the OIS covering measure~\cite{arbelaez2011contour} as the evaluation metric in Table \ref{tb:voc12_seg}. This metric is widely used for evaluating the generic segmentation methods and takes different categories into consideration. The FrontEnd (FE) represents to the method which only adopts the semantic labeling and extracts the single connected components as the instances. Combining FE with LLAM achieves higher OIS covering score than combining FE with NCuts does.

%Secondly, we use the standard instance segmentation measure and also compare with proposal-and-detection based models. For AR$@$10, LLAM achieves the comparable result with proposal-based methods and outperforms the NCuts model. Since LLAM is proposal-free, it doesn't generate hundreds of bouding boxes, leading to a relatively lower score on AR$@$100. As our model is the first step in proposal-free model for instance segmentation, our results are very promising.

\begin{table}[!htp] \small
\begin{center}
\begin{tabular} {c|c|c}
Method & AR$@$10 ($\%$) & AR$@$100 ($\%$) \\
\hline
{\bf Proposal-and-detection based} & \\
SS \cite{hariharan2014simultaneous} & $7.0$ & $23.5$ \\
MCG\cite{agrawal2014analyzing}  & $18.9$ & $36.8$ \\
DeepMask \cite{pinheiro2015learning} & $31.2$ & $42.9$ \\
MNC \cite{dai2015instance} & $33.4$ & $48.5$ \\
InstanceFCN \cite{dai2016instance} & $38.9$ & $49.7$ \\
\hline
\hline
{\bf Segmentation based}  & \\
Direct labeling (FE \cite{CP2015Semantic}) & $28.6$ &  $28.6$ \\
%FE+ground-truth-clustering& 37.4 & 39.3 \\
%FE+ground-truth-100& 37.9 & 39.8 \\
%\textcolor{mygray}{FE+ground-truth} & \textcolor{mygray}{38.2} & \textcolor{mygray}{39.9} \\
%FE+ground-truth-300& 37.2 & 39.3 \\
\hline
FE+Method 1 (ours) & $32.7$ &  $34.8$ \\
FE+Method 2 (ours) & $33.0$ &  $33.1$ \\
FE+Method 3 (ours) & $38.8$ & $39.2$

\end{tabular}
\caption{\footnotesize Comparison on the PASCAL VOC instance segmentation validation set based on AR@N. Models are grouped into proposal-and-detection based and segmentation based. FrontEnd (FE) refers to extracting connected components directly from semantic labeling prediction. For the transformation methods, Method 1 refers to pixel-based affinity mapping, Method 2 refers to superpixel-based affinity learning, and Method 3 refers to boundary-based componenet segmentaion. Our method is proposal-free and hence doesn't generate hundreds of proposals; this leads to relatively lower measure on AR@100. Our result with AR@10 measure that is fully segmentation based is comparable with other approaches that engage sliding windows for object detection. We do not find results with AR measures reported in PFN  \cite{liang2015proposal} and MPA \cite{MPA2016}. }
\label{tb:voc12_ar}
\end{center}
\vspace{-4mm}
\end{table}

\begin{figure*}[!htp]
\begin{center}
\begin{tabular} {c}
\includegraphics[width=0.8\linewidth]{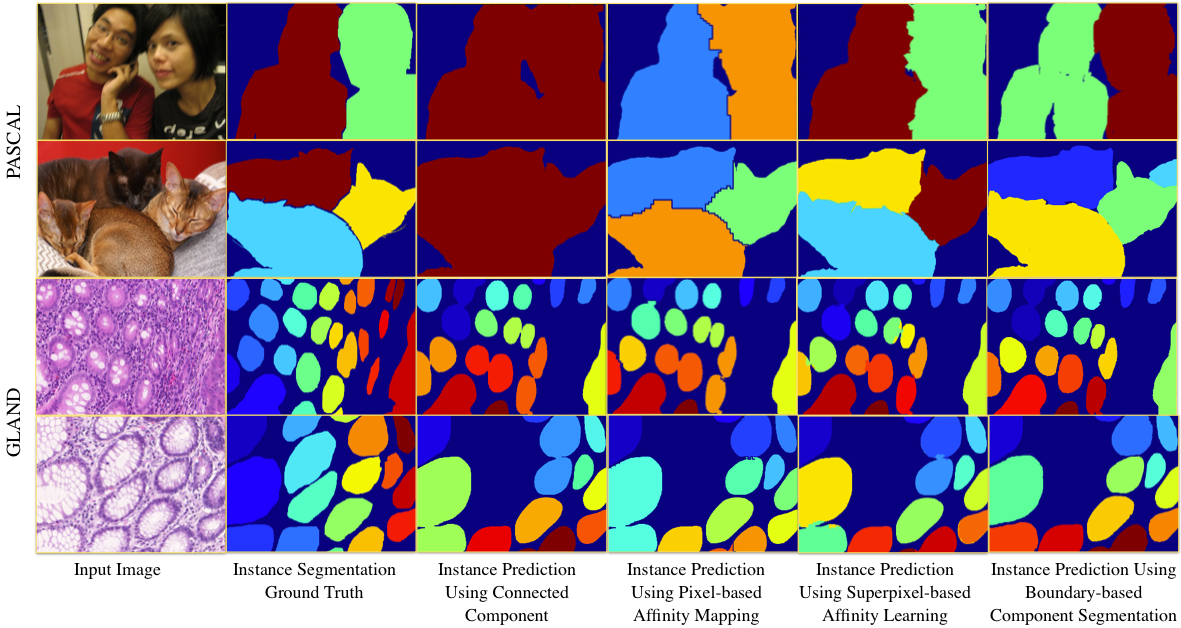} 
\end{tabular}
\caption{\footnotesize Example results of the instance segmentation on PASCAL VOC 2012 dataset \cite{pascal-voc-2012} and gland segmentation dataset \cite{miccai2015gland}. The first, second, third, fourth, fifth, and sixth column respectively shows the input image, its ground truth instance labeling, instance prediction from connected component of semantic segmentation, method 1 (instance prediction using pixel-based affinity mapping), method 2 (instance prediction using superpixel-based affinity learning) and method 3 (boundary-based component segmentation).}
\label{fig:gland_results}
\end{center}
\end{figure*}

Example instance segmentation results are shown in Figure \ref{fig:gland_results}. We observe that boundary-based component segmentation (Method 3) is more accurate to locate the object boundaries. For example, in the second example, boundary-based method can even fix the mistaken semantic segmentation results by separating the redundant part of the cat ear. However, affinity based methods may fail to identify the accurate separation of object instances in some cases.
% We also show a few instance segmentation results from our method 

%\subsection{COCO}
%The Microsoft COCO dataset~\cite{lin2014microsoft} is a recent challenging dataset for instance segmentation, which consists of more than 80k training images and more than 30k validation images. As the dataset does not provide image labeling, we created all the image labelling according to the given ground truth and computed the affinity map based on the instance map which are prodocued by ourselves. Similar to the experiment setting for the Pascal VOC 2012 dataset, we train our labeling map using the dilation front-end~\cite{yu2015multi} model and train another FCN for the affinity map model. We evaluate our performance on the first 1000 images from COCO validation set. Our LLAM model achieves relatively lower score than other proposal-and-detection based models.

\begin{table*}[!htp] \small
\newcolumntype{C}[1]{>{\centering\let\newline\\\arraybackslash\hspace{0pt}}m{#1}}
\begin{center}
\begin{tabular} {C{4cm}|C{1.5cm}|C{1.5cm}|C{1.5cm}|C{1.5cm}|C{1.5cm}|C{1.5cm}}
\multirow{2}{*}{Method} & \multicolumn{2}{c|}{F1 $\uparrow$} &  \multicolumn{2}{c|}{Object Dice $\uparrow$}  &  \multicolumn{2}{c}{Object Hausdorff $\downarrow$}  \\
%\cline{2-7} 
 & Part A & Part B & Part A & Part B & Part A & Part B \\
\hline
{\bf Proposal-and-detection based} &&&&&& \\

SDS  \cite{hariharan2014simultaneous}
&$0.545$&$0.322$&$0.647$&$0.495$&$116.833$&$229.853$\\
\hline

HyperColumn \cite{hariharan2015hypercolumn}	&$0.852$&$0.691$&$0.742$&$0.653$&$119.441$&$190.384$\\
\hline

MNC \cite{dai2015instance} & $0.856$&$0.701$&$0.793$&	$0.705$&$85.208$& $190.323$ \\
\hline

Frerburg2	\cite{ronneberger2015u}
&$0.870 $ &$0.695$	&$0.876$	&$0.786$	&$57.093$	&$148.463$\\
\hline

Xu et al. \cite{xu2016gland}	&$0.893$&$0.843$&$0.908$&$0.833$&$44.129$&$116.821$\\

\hline
\hline

{\bf Segmentation based} &&&&&& \\

CUMedVision2 \cite{chen2016dcan} &$0.912$	&$0.716$	&$0.897$	&$0.781$	&$45.418$	& $160.347$ \\
\hline

Direct labeling (FE \cite{CP2015Semantic})   &$0.844$	&$0.799$	&$0.873$	&$0.815$	&$63.321$	&$124.965$\\
\hline

FE+Method 1 (ours)	&$0.860$	&$0.813$	&$0.874$	&$0.828$	&$61.829$	&$110.865$\\
\hline

FE+Method 2 (ours)	&$0.874$	&$0.811$	&$0.879$	&$0.823$	&$60.881$	&$104.923$\\
\hline

FE+Method 3 (ours)	&$0.897$	&$0.820$	&$0.893$	&$0.827$	&$49.385$	&$106.978$\\

\end{tabular}
\caption{\footnotesize Comparison on the gland instance segmentation dataset based on the challenge measure \cite{miccai2015gland}. We report results on Part A and Part B. $\uparrow$ indicates a better performance if value is higher and $\downarrow$ indicates a better performance if the value is lower. FrontEnd (FE) refers to extracting connected components directly from semantic labeling prediction. Method 1 is pixel-based affinity mapping, Method 2 is superpixel-based affinity learning, and Method 3 is boundary-based componenet segmentaion. An object detection based method, MNC \cite{dai2015instance}, produces much worse result than ours, for the reason discussed before: detection-based methods assume rectangular objects and are hard to deal with objects of non-rigid shape. The results of our framework is on par with state-of-the-art performance on this dataset. Xu et al. \cite{xu2016gland} is much more complex than ours (consists of multiple modules with both semantic labeling and object detection) but reports performance that is very similar to ours.}
\label{tb:gland_results}
\end{center}
\end{table*}

% FrontEnd (FE) refers to extracting connected components directly from semantic labeling prediction. 

\subsection{MICCAI 2015 Gland Segmentation Dataset \label{sec:gland}}
The MICCAI 2015 Gland Segmentation Challenge dataset \cite{miccai2015gland} consists of 165 gland instance segmentation images, including 85 training images and 80 testing images. This dataset is texture-centric. We follow the data augmentation strategy in \cite{xu2016gland}, which includes horizontal flipping, rotation, sinusoidal transformation, pin cushion transformation and shear transformation.

We use three standard metrics from this challenge for evaluation: F1 measures the detection performance, ObjectDice measures the segmentation performance, and ObjectHausdorff measures the shape similarity. The results are summarized in Table \ref{tb:gland_results}. We observe that our proposed framework achieves better performance than detection-based methods \cite{hariharan2014simultaneous,hariharan2015hypercolumn,dai2015instance}. It is because that the shape of glands have large deformation. So they are hard to be bounded by rectangular boxes, which is the strong assumption from detection-based methods. 
Among three transformation methods, edge-based component segmentation method (Method 3) achieves the best performance, which is on par with state-of-the-art methods \cite{xu2016gland,chen2016dcan} on this dataset. 
Xu et al. method [38] is much more complex than ours (consists of multiple modules with both semantic labeling and object detection) but achieves similar performance.
Example instance segmentation results are shown in Fig. \ref{fig:gland_results}.

\section{Conclusions}
%\vspace{-0.13cm}
We have proposed an object detection free instance segmentation approach with three alternative methods when performing instance labeling transformations. We show competitive results on both PASCAL VOC 2012 and the gland segmentation datasets. Our methods have the desirable properties of being simple, transparent, easy to train, and fast to compute; it works well on both object-centric and texture-centric domains, which existing methods have not shown.

\noindent
{\bf Acknowledgments}
We would like to thank Saining Xie and Shuai Tang for valuable discussions.
This work is supported by NSF IIS-1216528 (IIS-1360566), NSF IIS-1618477, and a Northrop Grumman Contextual Robotics grant. We are grateful for the generous donation of the GPUs by NVIDIA.

{\small
\bibliographystyle{ieee}
\bibliography{main}
}

\end{document}